\documentclass{article}

\usepackage{multirow}
\usepackage{subfigure}
\usepackage[utf8]{inputenc}
\usepackage{amsmath}
\usepackage{hyperref}
\usepackage{url}
\usepackage{amssymb}
\usepackage{wrapfig}
\usepackage[overload]{empheq}
\usepackage{cleveref} 
\usepackage{CJK}
\usepackage{indentfirst}
\usepackage{mathtools}
\usepackage{amsmath}
\usepackage{amssymb}
\usepackage{mathtools}
\usepackage{amsthm}
\usepackage{comment}
\usepackage{color}
\usepackage{algorithm}
\usepackage{algorithmic}
\usepackage{braket}
\theoremstyle{plain}
\usepackage{microtype}
\usepackage{graphicx}
\usepackage{booktabs} 

\usepackage{hyperref}

\usepackage[accepted]{icml2024}

\usepackage{amsmath}
\usepackage{amssymb}
\usepackage{mathtools}
\usepackage{amsthm}
\usepackage{algorithmic}
\usepackage{algorithm}
\usepackage{caption}
\usepackage{stfloats}
\theoremstyle{plain}

\theoremstyle{definition}

\theoremstyle{remark}

\usepackage[textsize=tiny]{todonotes}

\icmltitlerunning{Harmonizing Generalization and Personalization in Federated Prompt Learning}

\begin{document}

\twocolumn[




\icmltitle{Harmonizing Generalization and Personalization \\in Federated Prompt Learning}




\icmlsetsymbol{correspond}{*}

\begin{icmlauthorlist}
\icmlauthor{Tianyu Cui}{sist}
\icmlauthor{Hongxia Li}{sist}
\icmlauthor{Jingya Wang}{sist}
\icmlauthor{Ye Shi}{sist,correspond}
\end{icmlauthorlist}

\icmlaffiliation{sist}{ShanghaiTech University}

\icmlcorrespondingauthor{Ye Shi}{shiye@shanghaitech.edu.cn}

\icmlkeywords{Machine Learning, ICML}

\vskip 0.3in
]



\printAffiliationsAndNotice{}  

\begin{abstract}

Federated Prompt Learning (FPL) incorporates large pre-trained Vision-Language models (VLM) into federated learning through prompt tuning. The transferable representations and remarkable generalization capacity of VLM make them highly compatible with the integration of federated learning. Addressing data heterogeneity in federated learning requires personalization, but excessive focus on it across clients could compromise the model's ability to generalize effectively. To preserve the impressive generalization capability of VLM, it is crucial to strike a balance between personalization and generalization in FPL. To tackle this challenge, we proposed \textbf{Fed}erated \textbf{P}rompt Learning with CLIP \textbf{G}eneralization and low-rank \textbf{P}ersonalization (FedPGP), which employs pre-trained CLIP to provide knowledge-guidance on the global prompt for improved generalization and incorporates a low-rank adaptation term to personalize the global prompt. Further, FedPGP integrates a prompt-wise contrastive loss to achieve knowledge guidance and personalized adaptation simultaneously, enabling a harmonious balance between personalization and generalization in FPL. We conduct extensive experiments on various datasets to explore base-to-novel generalization in both category-level and domain-level scenarios with heterogeneous data, showing the superiority of FedPGP in balancing generalization and personalization. Code is available at https://github.com/TianyuCuiOvO/FedPGP.
\end{abstract}

\section{Introduction}
Federated Learning \cite{mcmahan2017communication} has been proposed as an efficient collaborative learning strategy, enabling clients to jointly train a global model while preserving data privacy. In this context, the ability of large pre-trained Vision-Language models (VLM) like CLIP \cite{radford2021learning} and ALIGN \cite{jia2021scaling} to learn transferable representations across downstream tasks makes them a natural fit for integration with federated learning. This collaborative approach not only harnesses the outstanding performance and generalization capabilities of pre-trained models but also ensures efficient and privacy-preserving global model training across multiple clients.
However, due to the millions of parameters in VLM, fine-tuning the entire model in federated learning leads to high communication costs and memory footprint issues. Prompt tuning addresses these challenges by adapting pre-trained models to diverse downstream tasks with a reduced parameter count, and its integration of federated learning has been explored in previous research \cite{zhao2022reduce, guo2023promptfl}. We term the combination of prompt-tuning and federated learning as Federated Prompt Learning (FPL) for simplicity. Currently, studies in FPL have not been thoroughly explored in terms of personalization and generalization. Methods derived from traditional federated learning studies fail to capture the multi-modality of VLM, which hinders a direct transfer of these methods into FPL. 

In federated learning, it is essential to account for the generalization capability to unseen domains or categories. {\color{black} However, existing studies in FL, like \cite{nguyen2022fedsr, liu2021feddg, zhang2021federated}, have struggled to achieve satisfactory results in evaluating generalization on target datasets in unseen domains.}
{\color{black} With the help of VLM which has strong generalization performance, this problem may be solved.
Unfortunately, the generalization issues in prompt-based VLM have been revealed in recent research \cite{khattak2023maple, khattak2023self}.}
For instance, CoOp \cite{zhou2022learning} struggles with generalizing to unseen categories within the same dataset due to overfitting, resulting in lower test accuracy on novel categories compared to the zero-shot CLIP baseline with a handcrafted prompt.
PromptSRC \cite{khattak2023self} addresses this issue with self-regularization constraints, maximizing the mutual agreement between prompted and frozen VLM features. However, the generalization of FPL is still an open challenge. FedTPG \cite{qiu2023text} takes a step toward exploring generalization by learning a unified prompt generation network among multiple clients but disregards the data heterogeneity. 

Data heterogeneity results in another challenge in federated learning, where the data distributions among clients are not independently and identically distributed (Non-IID). This leads to discrepancies between local and global optimization objectives, making it difficult for a single global prompt to adapt to the varied local distributions.
In the endeavor to learn personalized models, pFedPrompt \cite{guo2023pfedprompt} incorporates personalized attention modules into FPL while learning a consensus among users through shared text prompts. 
{\color{black} Nevertheless, if we employ strong personalized techniques to fully adapt the prompts to local distributions, it may lead to the loss of inherent generalization in VLM.}
This raises the question we aim to explore: 

\textit{How can we strike a balance between generalization and personalization in FPL?}

To overcome the problems outlined above, we proposed \textbf{Fed}erated \textbf{P}rompt Learning with CLIP \textbf{G}eneralization
and low-rank \textbf{P}ersonalization (FedPGP), an effective method that reaches a balance between personalization and generalization. 
In FedPGP, each client learns a personalized prompt, combining a global prompt and an adaptation term to accommodate heterogeneous local distributions. The incorporation of the adaptation term allows fine-tuning of the global prompt for specific client needs. 
To enhance prompt generalization, we incorporate category-agnostic knowledge from CLIP, aligning the global prompt in each client towards a unified direction.

To balance generalization and personalization, we utilize a low-rank decomposition for the adaptation term,
ensuring robust generalization capabilities in comparison to a full-rank term.
To enable personalized prompts better access to client-specific knowledge, we aim to separate representations of global and personalized prompts for better personalization.
Considering this and the knowledge-guidance from CLIP to global prompt, we introduce an additional contrastive loss into the optimization objective to further balance personalization and generalization of our FedPGP. This involves treating global prompt representations with personalized ones as negative pairs for personalization, while simultaneously treating them as positive pairs with the representation of handcrafted prompt of CLIP for generalization. 

Our main contributions are summarized as follows: 
\begin{itemize}
    \item We are the first to consider both personalization and generalization in federated prompt learning. {\color{black} We aim to learn a personalized prompt for each client in heterogeneous federated scenarios while preserving the remarkable generalization capacity in Visual-Language models, leading to the balance of generalization and personalization.}

    \item We propose FedPGP, utilizing low-rank decomposition adaptation to flexibly adjust the global prompt to heterogeneous local distributions, which prevents overfitting on local datasets. 
    Additionally, we integrate an extra contrastive loss, treating representations of global and personalized prompts as negative pairs and representations of global and handcrafted prompts as positive pairs.
    
    

    \item  We conduct extensive experiments on widely adopted datasets to investigate the base-to-novel generalization of FedPGP on both category-level and domain-level in the case of heterogeneous data. Our comparative experimental results demonstrate FedPGP's superiority in harmonizing generalization and personalization.

\end{itemize}

\section{Related Work}
\textbf{Federated Learning}
This subsection mainly introduces the research on personalization and generalization in federated learning.
Personalized federated learning (PFL) algorithms, rather than creating a universal model for all clients, tackle data heterogeneity by learning customized models for each client. Various strategies for achieving PFL have been suggested in previous studies. Some existing methods combine global model optimization with additional local model customization involving local fine-tuning \cite{wang2019federated,mansour2020three,tan2022towards}, regularization \cite{li2020federated,li2021ditto,t2020personalized}, parameter decomposition \cite{jeong2022factorized,hyeon2021fedpara,arivazhagan2019federated}, parameter generation \cite{shamsian2021personalized, ma2022layer, li2023fedtp}, and clustering methods for client grouping \cite{huang2021personalized,zhang2020personalized,sattler2020clustered,zhang2022federated,cao2023knowledge,cai2024fed}. The theoretical significance of PFL was pointed out by \cite{huang2023understanding}. To be specific, 
FedPer \cite{arivazhagan2019federated}, FedBABU \cite{oh2021fedbabu}, and FedRep \cite{collins2021exploiting} share the base layers while learning personalized classifier heads locally. FedBN \cite{li2021fedbn} uses local batch normalization to alleviate feature shift before averaging models. FedRod \cite{chen2021bridging} proposed learning a global predictor for generic FL and a local predictor for personalized FL. 

Many existing studies in federated learning commonly assume the test dataset is a subset of the client dataset. However, a research gap exists for scenarios where the target dataset (i.e., the test dataset) is not included in the training process. This scenario is also referred to as domain generalization in centralized machine learning. 
FedSR \cite{nguyen2022fedsr} employs regularization techniques for simplified data representation, intending to achieve improved generalization capabilities. ELCFS \cite{liu2021feddg} tackled federated domain generalization with continuous frequency space interpolation and the boundary-oriented episodic learning scheme. FedADG \cite{zhang2021federated} utilizes federated adversarial learning for dynamic universal feature representation. Unlike traditional federated learning approaches, we are the first to consider both personalization and generalization in the context of FPL.

\textbf{Federated Prompt Learning}
Prompt tuning is a technique employed to adapt pre-trained models to diverse downstream tasks. For instance, CoOp \cite{zhou2022learning} uses tunable text prompts to replace the fixed template in CLIP, and CoCoOp \cite{zhou2022conditional} utilizes image feature to instruct the optimization of the soft text prompt. 
ProGrad \cite{zhu2023prompt} selectively updates prompts based on aligned gradients with general knowledge to prevent forgetting essential information from VLMs.
Some works also incorporate prompt tuning into federated learning. PromptFL \cite{guo2023promptfl} introduced prompt learning into Federated Learning. 
FedPR \cite{feng2023learning} focuses on learning federated visual prompts within the null space of the global prompt for MRI reconstruction.
pFedPG \cite{yang2023efficient} employs a client-specific prompt generator on the server side for personalized prompts, while FedTPG \cite{qiu2023text} also trains a global prompt generation network to enhance generalization. 
pFedprompt \cite{guo2023pfedprompt} maintains a non-parametric personalized attention module for each client to generate local personalized spatial visual features. FedOTP \cite{li2024global} employs unbalanced Optimal Transport to promote the collaboration between global and local prompts across heterogeneous clients. Designed for domain discrepancy, FedAPT \cite{wei2023dual} unlocks specific domain knowledge for each test sample to provide personalized prompts, and Fed-DPT \cite{su2022cross} applies both visual and textual prompt tuning to facilitate domain adaptation over decentralized data. However, these methods overlook the aspect of generalization.
We propose FedPGP, a framework that effectively balances personalization and generalization in FPL.



\textbf{Contrastive learning}
Contrastive learning methodologies have gained significant attention by consistently attaining state-of-the-art outcomes results in the field of visual representation learning \cite{chen2020simple,chen2020improved,xie2022simmim}. 
The fundamental principle behind contrastive learning is to minimize the distance between representations generated from diverse augmentations of the same image (positive pairs) while simultaneously maximizing the distance between representations obtained from augmented views of different images (negative pairs).
A proportion of research \cite{tan2022federated,li2021model,mu2023fedproc} combines contrastive learning with federated learning, which improves the local training process and achieves higher model effectiveness. FedPCL \cite{tan2022federated} employs prototype-wise contrastive learning for client-specific representations, promoting alignment with global and local prototypes to enhance knowledge sharing.
MOON \cite{li2021model} minimizes the distance between local and global model representations while increasing the distance from the previous local model's representation.
Different from model-wise of MOON and prototype-wise of FedPCL, our FedPGP introduces a novel prompt-wise contrastive methodology. In addition, unlike previous work focusing on aligning the global and local components, FedPGP treats representations of global and personalized prompts as negative pairs and representations of global and handcrafted prompts as positive pairs.

\section{Proposed Method}
\begin{figure*}[h]
\begin{center}
\includegraphics[width=0.95\linewidth]{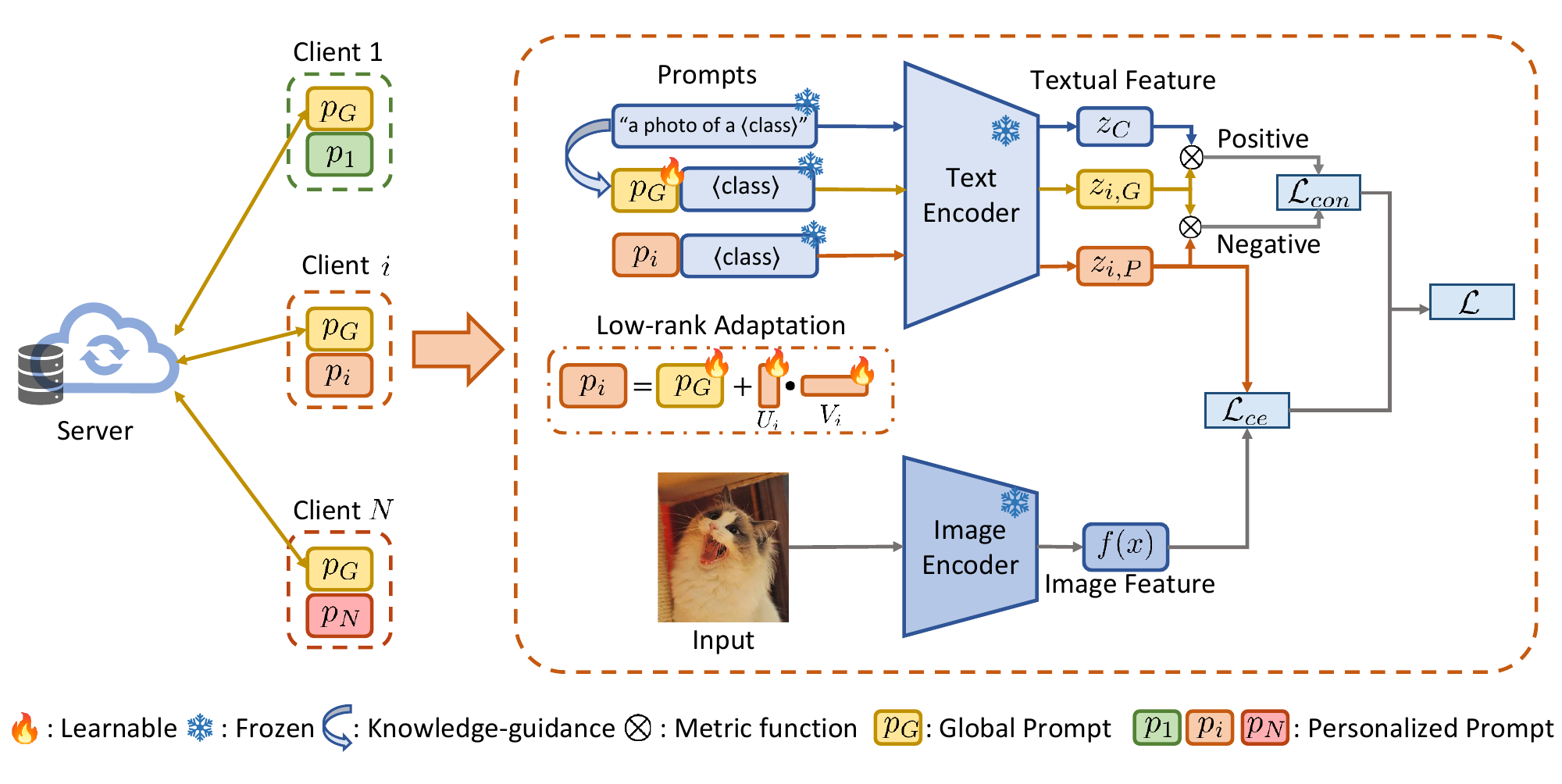}
\end{center}
\caption{Pipeline of FedPGP. On the left, clients send global prompts to the server for aggregation while retaining adaptation term locally. The right shows the workflow of CLIP knowledge-guidance and low-rank adaptation with an additional contrastive loss to balance generalization and personalization.} 
\label{pipeline}
\end{figure*}

In this section, we delve into the details of our proposed FedPGP, illustrated in Figure \ref{pipeline}. 
FedPGP leverages CLIP knowledge-guidance and low-rank adaptation with an additional contrastive loss to balance generalization and personalization.

\subsection{Preliminaries of Prompt Learning}
Prompt learning methods \cite{zhou2022learning, zhou2022conditional} offer an efficient approach to adapting pre-trained models like CLIP to downstream tasks by training a part of the parameters in the prompt. Unlike the zero-shot transfer that utilizes a fixed word embedding $W = \{w_1,w_2,...,w_l \}$ mapped from a hand-crafted prompt (e.g., ``a photo of a ⟨label⟩"), prompt learning replaces a set of $M$ continuous context vectors $p = \{p_1,...,p_M\}  \in \mathbb{R}^{d\times k}$ as the learnable prompt. Specifically, we use $p = \{p_1,...,p_M\}$ to replace $\{w_2,...,w_{M+1}\}$ to be consistent with previous methods. Then the textual prompt of $k$ class can be reformulated as $P_k = \{w_1,p_1,...,p_M,w_{M+2},...,w_l\}$ and is fed into pre-trained text encoder $g(\cdot)$. Denote image encoder as $f(\cdot)$, the prediction probability for each category of input image $x$ is computed through matching scores: 
\begin{equation}
    p(\hat{y}=k|x) = \frac{\exp(\text{sim}(f(x),g(P_k))/\tau)}{\sum_
    {c=1}^{K} \exp(\text{sim}(f(x),g(P_c))/\tau)},
\end{equation}
where $\text{sim}(\cdot,\cdot)$ denotes a metric function (e.g., cosine similarity), $\hat{y}$ denotes the predicted label, $K$ denotes the number of classes, and $\tau$ denotes the temperature of Softmax. Then we optimize the learnable prompt by cross-entropy loss:
\begin{equation}
    \mathcal{L}_{ce} = - \frac{1}{|\mathcal{D}|}\sum_{(x,y) \in \mathcal{D}}\sum_k y\log p(\hat{y}=k|x) ,
\end{equation}
where $y$ denotes the one-hot ground-truth annotation.

\subsection{Federated Prompt Learning}
Suppose there are $N$ clients and a central server. Each client $i$ holds local dataset $D_i$ with $n_i$ samples and $D = \{D_1,D_2,...,D_N\}$ represents the total dataset where each dataset is derived from a distinct data distribution $\mathcal{D}_i$. Each client is equipped with a pre-trained CLIP model and a prompt learner in our federated learning setup. Let $C_t$ represent the set of selected clients participating in communication round $t$. For each communication round $t$, the selected clients initialize the global prompt with $p_G^{t-1}$ and perform local training $p_i^t$ through cross-entropy loss $\mathcal{L}_{ce}$ for $E$ local epoch, at $e$ local epoch the update of global prompt is:
\begin{equation}
    p_{G,i}^{t,e} = p_{G,i}^{t,e-1} - \eta \nabla \mathcal{L}_{ce}(p_{G,i}^{t,e-1}).
\end{equation}
After $E$ local epoch training, each client in $C_t$ uploads the global prompt $p_{G,i}^{t,E}$ to the server for aggregation:
\begin{equation}
    p_G^t = \sum_{i \in C_t} \frac{n_i}{\sum_{j \in C_t} n_j} p_{G,i}^{t,E}.
\end{equation}
The global prompt is aggregated in the context of federated learning, carrying the unique characteristics learned from other clients. The optimization objective of FPL can be formulated as:
\begin{equation}
    \min_{p_G} \sum_{i=1}^N \frac{n_i}{\sum_j n_j} \mathcal{L}_{ce}^{D_i}(p_G),
\end{equation}
where $\mathcal{L}_{ce}^{D_i}(p_G)$ represents the cross-entropy loss on dataset $D_i$ of client $i$.

\subsection{Generalization and Personalization for FPL}
\label{3.3}
Previous research has discovered that prompted visual-language models, such as CoOp, overfit to the base classes and cannot generalize to the unseen class observed during training \cite{zhou2022conditional,zhu2023prompt,ma2023understanding}.
This phenomenon of overfitting to base classes implies that the prompt fails to capture more generalized elements that are crucial for recognizing a wider range of scenarios. On the contrary, the manually designed prompts adopted by the zero-shot CLIP are relatively generalizable. The problem of generalization in prompt vision-language models remains unresolved in FPL. The objective of the client's local training can be formulated as:
\begin{equation}
    \mathcal{L}_{ce}^{D_i}(p_{G,i}^{t,e}) =- \frac{1}{|\mathcal{D}_i|}\sum_{(x,y) \in \mathcal{D}_i}\sum_k y\log p(\hat{y}=k|x).
\end{equation}
$\mathcal{L}_{ce}^{D_i}$ is to optimize the prompts for the client-specific task. 
Despite aggregating prompts in federated learning, overfitting to client-specific tasks remains a challenge. Leveraging the remarkable generalization capabilities of CLIP, FedPGP utilizes knowledge from CLIP to guide the global prompt to enhance generalization. Specifically, we obtain the representations of the handcrafted prompt $g(p_{C})$ of CLIP and the global prompt $g(p_G)$ and align them through a metric function, such as cosine similarity. This knowledge-guidance from CLIP promotes the preservation of category-agnostic information within learnable global prompts, contributing to improve model generalization.


Due to data heterogeneity, it is difficult for a single global prompt to adapt to diverse local distributions. Different from tuning model parameters in traditional federated learning, FPL involves frozen client models and learnable prompts, leading to a distinct approach in adapting global prompt to local distributions.
In FedPGP, the adaptation of global prompt $p_G$ to the client-specific prompt $p_i$ is achieved by introducing an additional adaptation term $\Delta p_i$:
\begin{equation}
    p_i = p_G + \Delta p_i,
\end{equation}
where $\Delta p_i \in \mathbb{R}^{d\times k}$ owns the same dimension of $p_G$. Then the objective of federated learning can be formulated as:
\begin{equation}
    \min_{p_G,\{\Delta p_i\}_{i=1}^N} \sum_{i=1}^N \frac{n_i}{\sum_j n_j} \mathcal{L}_{ce}^{D_i}(p_G + \Delta p_i).
\end{equation}

\subsection{Balance FPL's Generalization and Personalization}
Previous research \cite{aghajanyan2020intrinsic} has shown that pre-trained language models with lower intrinsic dimensions tend to exhibit better evaluation accuracy and lower relative generalization gaps across various tasks. Inspired by this, we propose that the prompt may also possess a low "intrinsic rank" during the adaptation process. To retain information derived from aggregation and knowledge-guidance of CLIP, our adaptation term is designed in a low-rank form instead of adding a full-rank term to overwrite the global prompt entirely. Specifically, the additional term is decomposed as:
\begin{equation}
    \Delta p_i = U_i V_i.
\end{equation}
We decompose $\Delta p_i$ into multiplication between two low-rank matrices $U_i \in \mathbb{R}^{d\times b}$ and $V_i \in \mathbb{R}^{b \times k}$, where $b$ denotes the bottleneck dimension of low-rank decomposition. Consequently, each client's personalized learnable prompt $p_i \in \mathbb{R}^{d\times k}$ can be reformulated as:
\begin{equation}
    p_i = p_G + \Delta p_i = p_G + U_i V_i,
\end{equation}
where $p_G \in \mathbb{R}^{d\times k}$ is the full-rank matrix and  $\Delta p_i$ is the low-rank component of personalized $p_i$. Consequently, the objective of FedPGP can be reformulated as:
\begin{equation}
    \min_{p_G,\{U_i,V_i\}_{i=1}^N} \sum_{i=1}^N \frac{n_i}{\sum_j n_j} \mathcal{L}_{ce}^{D_i}(p_G + U_iV_i).
\end{equation}
Employing the low-rank adaptation, FedPGP introduces personalization while preserving generalizability, striking a balance between the model's ability to generalize and personalize. 
Moreover, our objective is to go beyond the consensus knowledge communicated by clients through the global prompt and instead offer them personalized knowledge. To enable personalized prompts better access to client-specific knowledge, we aim to increase dissimilarity between representations of global and personalized prompts for better personalization. 


Summarizing the target we mentioned, our objective is 1) to bring close the representations of the handcrafted prompt $g(p_{C})$ of CLIP and the global prompt $g(p_G)$, and 2) to create a clear distinction between the representations of the global prompt $g(P_G)$ and personalized prompt $g(P_i)$. Building upon the above analysis, 
We consider the global prompt representations $z_G$ with handcrafted prompt representation $z_{C}$ as positive pairs, while simultaneously treating them as negative pairs with personalized prompt representations $z_i$.
Consequently, we design an additional contrastive loss $\mathcal{L}_{con}$ for FedPGP to balance generalization and personalization:
\begin{equation}
\label{con_loss}
    \mathcal{L}_{con}\! =\! -\! \log \frac{\exp( \text{sim}(z_G,\!z_{C})/ \tau)}{\exp( \text{sim}(z_G,\!z_{C})\!/\!\tau) \! + \!\exp( \text{sim}(z_G,\!z_i)\!/\!\tau)},
\end{equation}
where $\text{sim}(\cdot,\cdot)$ denotes a metric function (e.g., cosine similarity).
The contrastive loss can guide the global prompt to gain complementary knowledge from pre-trained CLIP representation and enable $\Delta p_i$ to learn personalized knowledge distinct from the global prompt.
Consequently, our overall training objective thus becomes:
\begin{equation}
    \mathcal{L} = \mathcal{L}_{ce} + \mu \mathcal{L}_{con},
\end{equation}
where $\mu \geq 0$ is a hyper-parameter. We offer comprehensive algorithmic details for FedPGP in Algorithm \ref{alg:algorithm}. For every communication round $t$, the selected clients locally train both the global prompt $p_{G,i}$ and the low-rank adaptation term $\Delta p_i$. After local training, the updated global prompt $p_{G,i}^{t,E}$ are sent to the server for aggregation, while the low-rank adaptation term $\Delta p_i$ is retained locally.

\begin{algorithm}[ht]
\caption{FedPGP}
\label{alg:algorithm}
\textbf{Input}: Communication rounds $T$, local epochs $R$, client number $N$, local dataset $D_i$ , sample numbers $m_i$, pre-trained CLIP model text encoder $g(\cdot)$ and image encoder $f(\cdot)$, class number $K$, learning rate $\eta$, the temperature of Softmax $\tau$, hyper-parameter$\mu$, and bottleneck number $b$.
\begin{algorithmic}[1] 
\STATE Initialize parameters $p_i^0 = p_G^0 + \Delta p_i^0$
\FOR{each communication rounds $t \in \{1,...,T\}$}
\STATE Sample client $C^t \in \{1,...,N\}$
\FOR{each client $i \in C^t$}
\STATE Initialize $p_G^{t,0} = p_G^{t-1}$,$p_i^{t,0} = p_G^{t,0}+ \Delta p_i^{t-1}$
\FOR{each local epoch $e \in \{1,...,E\}$}
\STATE Sample a mini-batch $B_i \in D_i$
\STATE Obtain the image feature $f(x) (x \in B_i)$ through image encoder $f(\cdot)$
\STATE Obtain the global text feature $g(P_G^{t,e})$ ,the personalized text feature $g(P_i^{t,e})$, the CLIP general text feature $g(P_C)$ through text encoder $g(\cdot)$
\STATE Calculate the cross-entropy loss $\mathcal{L}_{ce}$ ,the contrastive loss $\mathcal{L}_{con}$ according to (\ref{con_loss}) and the optimization objective $\mathcal{L} = \mathcal{L}_{ce} + \mu \mathcal{L}_{con}$
\STATE Update prompts $p_{G,i}^{t,e} \leftarrow  p^{t,e-1} _G  - \eta \nabla \mathcal{L}_{D_i}$
\ENDFOR
\ENDFOR
\STATE Aggregate and calculate the global prompt $p_G^t = \sum_{i\in C_t} \frac{n_i}{\sum_{j \in C_t} n_j}p_{G,i}^{t,E}$
\ENDFOR
\STATE \textbf{return} $p_i = p_G + \Delta p_i$
\end{algorithmic}
\end{algorithm}

\section{Experiments}
In this section, we conduct extensive experiments aiming at evaluating the generalization and personalization capability of FedPGP in scenarios of heterogeneous data distribution.

\subsection{Experimental Setup}
\noindent \textbf{Datasets and Data Heterogeneity.} 
Following previous research \cite{guo2023pfedprompt,guo2023promptfl}, we selected five datasets to investigate base-to-novel class generalization ability: OxfordPets \cite{parkhi2012cats}, Flowers102 \cite{nilsback2008automated}, DTD \cite{cimpoi2014describing}, Caltech101 \cite{fei2004learning}, Food101 \cite{bossard2014food}. We equally split the datasets into base and novel classes and utilized the pathological setting by assigning a specific number of non-overlapping base classes to each client. 
Each client model is trained on their local classes and evaluated on both local classes, base classes (classes seen on other clients), and novel classes (unseen in the whole training process).
For domain generalization, we evaluate FedPGP on two datasets with multi-domains: DomainNet \cite{peng2019moment} with six domains and Office-Caltech10 \cite{gong2012geodesic} with four domains.
Similar to previous research \cite{nguyen2022fedsr,zhang2021federated}, we utilize the leave-one-domain-out validation strategy. Each client participating in the federated learning system is assigned data from one of the distinct domains. We pick one domain to serve as the target domain and use the rest as source domains. Each client possesses a distinct source domain for training and then tests its model generalization ability on the whole target domain.

For evaluation of personalization, beyond the datasets used in base-to-novel class generalization, we employed two additional benchmark datasets: CIFAR-10 \cite{krizhevsky2010cifar} and CIFAR-100 \cite{krizhevsky2009learning}. We applied the Dirichlet Distribution, as in previous work where the datasets were partitioned randomly among clients using a symmetric Dirichlet distribution. Besides, we employ the Pathlogicacl setting the same as in base-to-novel class generalization with non-overlapping classes across clients. 
The Appendix Section \ref{dataset_setup} contains comprehensive information regarding each dataset and provides additional details about the Non-IID settings.

\begin{table*}[ht!] 
    \caption{Accuracy comparison ($\%$) on clients’ local classes and Base-to-novel generalization.}
    \begin{minipage}{0.5\linewidth} \label{basetonew}
         \centering
         \renewcommand\arraystretch{1.1}
        \caption*{(a) Average over 5 datasets.}
        \resizebox{\linewidth}{!}{
        \begin{tabular}{l| ccc|c}
        \toprule 
        Methods & Local & Base & Novel & HM \\
        \midrule
        CLIP \cite{radford2021learning}  & 79.18  & 79.83  & \textbf{83.25}  & 80.72 \\ 
        CoOp \cite{zhou2022learning}  & 94.28  & 69.40  & 73.16  & 77.55 \\ 
        PromptFL \cite{guo2023promptfl}  & 90.00  & 85.65  & 78.53  & 84.46 \\ 
        Prompt+Prox \cite{li2020federated} &89.84 &	85.04 &	77.40 &	83.78  \\
        FedMaPLe & 90.81   & 84.90     &  81.49   &  85.56   \\
        FedCoCoOp & 90.01    & 85.08   &   81.4  & 85.35    \\
        FedOTP \cite{li2024global}& \textbf{98.16} &59.73  & 71.08 & 73.17 \\
        \midrule
        FedPGP & 95.67  & \textbf{85.69}  & 81.75  & \textbf{87.33} \\
        \bottomrule
        \end{tabular}
        }

    \end{minipage}
    ~
    \begin{minipage}{0.5\linewidth}
        \centering
        \renewcommand\arraystretch{1.1}
        \caption*{(b) OxfordPets.}
        \resizebox{\linewidth}{!}{
        \begin{tabular}{l| ccc|c}
        \toprule 
        Methods & Local & Base & Novel & HM \\
        \midrule
        CLIP \cite{radford2021learning} & 89.34 & 89.31 & 96.86 & 91.70   \\ 
        CoOp \cite{zhou2022learning} & 95.33 & 82.51 & 92.92 & 89.90   \\ 
        PromptFL \cite{guo2023promptfl} & 95.12 & 95.16 & 91.89 & 94.03   \\
        Prompt+Prox \cite{li2020federated} &95.95 	&95.24 &	91.25 &	94.10 \\
        FedMaPLe &    93.75&  95.53    &  \textbf{97.45}   &  95.55     \\
        FedCoCoOp &  96.02   & 96.01   &  97.25  & 96.42  \\
        FedOTP \cite{li2024global}& \textbf{99.93}  & 63.92 & 80.56 & 78.81 \\
        \midrule
        FedPGP & 96.65 & \textbf{95.87} & 97.33 & \textbf{96.61} \\ 
        \bottomrule
        \end{tabular}
        }
    \end{minipage}
    ~
    \begin{minipage}{0.5\linewidth}
        \centering
        \renewcommand\arraystretch{1.1}
        \caption*{(c) Flowers102.}
        \resizebox{\linewidth}{!}{
        \begin{tabular}{l| ccc|c}
        \toprule 
        Methods & Local & Base & Novel & HM \\
        \midrule
         CLIP \cite{radford2021learning} & 67.69 & 68.85 & \textbf{77.23} & 71.01   \\ 
        CoOp \cite{zhou2022learning} & 96.39 & 55.91 & 64.47 & 68.54   \\ 
        PromptFL \cite{guo2023promptfl} & 94.32 & 76.19 & 70.1 & 78.96   \\ 
        Prompt+Prox \cite{li2020federated} & 92.73 & 73.06 &	66.09 &	75.75 \\
        FedMaPLe & 94.89   &  77.49    &  70.46   &  79.71   \\
        FedCoCoOp &  94.57   &  77.88  &  74.39   &  81.4   \\
        FedOTP \cite{li2024global}&99.63  &  44.16  & 56.57 &59.57 \\
        \midrule
        FedPGP & \textbf{99.68} & \textbf{78.48} & 75.11 & \textbf{83.13}   \\ 
        \bottomrule
        \end{tabular}
        }
    \end{minipage}
    ~
    \begin{minipage}{0.5\linewidth}
        \centering
        \renewcommand\arraystretch{1.1}
        \caption*{(d) DTD.}
        \resizebox{\linewidth}{!}{
        \begin{tabular}{l| ccc|c}
        \toprule 
        Methods & Local & Base & Novel & HM \\
        \midrule
       CLIP \cite{radford2021learning} & 53.79 & 54.62 & \textbf{58.20} & 55.47   \\
        CoOp \cite{zhou2022learning} & 86.38 & 39.2 & 37.65 & 47.13   \\
        PromptFL \cite{guo2023promptfl} & 72.71 & 71.41 & 49.28 & 62.44   \\ 
        Prompt+Prox \cite{li2020federated} & 74.07 &	\textbf{71.84} &	50.20 &	63.37 \\
        FedMaPLe &  78.37  &  65.35    &  55.85   &  65.26   \\
        FedCoCoOp &  72.61   &  68.20  &  54.4   &  64.08   \\
        FedOTP \cite{li2024global}& \textbf{94.91} & 43.87& 43.68 & 53.36\\
        \midrule
        FedPGP & 89.07 & 69.65 & 55.25 & \textbf{68.15}   \\ 
        \bottomrule
        \end{tabular}
        }
    \end{minipage} 
    ~
    \begin{minipage}{0.5\linewidth}
        \centering
        \renewcommand\arraystretch{1.1}
        \caption*{(e) Caltech101.}
        \resizebox{\linewidth}{!}{
        \begin{tabular}{l| ccc|c}
        \toprule 
        Methods & Local & Base & Novel & HM \\
        \midrule
        CLIP \cite{radford2021learning} & 95.72 & 96.96 & 93.99 & 95.54   \\ 
        CoOp \cite{zhou2022learning} & 99.39 & 86.37 & 86.12 & 90.22   \\ 
        PromptFL \cite{guo2023promptfl} & 97.04 & 97.27 & 92.79 & 95.65   \\ 
        Prompt+Prox \cite{li2020federated} & 96.76 &	\textbf{97.34} &	91.99 &	95.30 \\
        FedMaPLe & 96.47   &   96.7   &  \textbf{94.32}   &  95.82   \\
        FedCoCoOp &  96.65   &  95.45  &  92.46   &  94.82   \\
        FedOTP \cite{li2024global}& \textbf{99.68} & 87.49 & 89.33& 91.86\\
        \midrule
        FedPGP & 99.46 & 96.09 & 93.62 & \textbf{96.33} \\ 
        \bottomrule
        \end{tabular}
        }
    \end{minipage} 
    ~
    \begin{minipage}{0.5\linewidth}
        \centering
        \renewcommand\arraystretch{1.1}
        \caption*{(f) Food101.}
        \resizebox{\linewidth}{!}{
        \begin{tabular}{l| ccc|c}
        \toprule 
        Methods & Local & Base & Novel & HM \\
        \midrule
        CLIP \cite{radford2021learning} & 89.38 & 89.39 & \textbf{89.98} & 89.58   \\ 
        CoOp \cite{zhou2022learning} & 93.92 & 82.99 & 84.62 & 86.92   \\ 
        PromptFL \cite{guo2023promptfl} & 90.79 & 88.22 & 88.6 & 89.19   \\
        Prompt+Prox \cite{li2020federated} & 89.68 &	87.72 &	87.49 &	88.29 \\
        FedMaPLe &  90.59  &  \textbf{89.43}    & 89.38    &  89.80   \\
        FedCoCoOp & 90.18    &  87.86  &   88.51  &  88.84   \\
        FedOTP \cite{li2024global}& \textbf{96.65} & 59.19 & 85.28 & 76.98\\
        \midrule
        FedPGP & 93.51 & 88.37 & 88.44 & \textbf{90.04}   \\ 
        \bottomrule
        \end{tabular}
        }
    \end{minipage} 
    \label{tab:my_label}
\end{table*}

\noindent \textbf{Baselines.} 
For generalization, we compare FedPGP with (i) Zero-shot CLIP \cite{radford2021learning} with hand-crafted text prompt template, e.g., ``a photo of a [class]" (ii) CoOp \cite{zhou2022learning} with learnable prompt vectors replacing hand-crafted text prompts trained on each client locally. (iii) PromptFL \cite{guo2023promptfl} with unified prompt vectors learned across clients via FedAvg \cite{mcmahan2017communication} collectively. For personalization, we consider (iv) pFedPrompt \cite{guo2023pfedprompt} which learns a unified prompt with personalized attention modules for each client and four baselines introduced in \cite{guo2023pfedprompt}, which are derived from traditional personalized federated learning techniques: (v) PromptFL+FT \cite{cheng2021fine}, (vi) Prompt+Per \cite{arivazhagan2019federated}, (vii) Prompt+Prox \cite{li2020federated} and Prompt+AMP \cite{huang2021personalized}.


\begin{table*}[!ht]
\renewcommand\arraystretch{1.12}
    \centering
    \caption{The average classification accuracy using leave-one-domain-out validation on Offica-Caltech10 and DomainNet.}
    \label{leave_one_domain_out}
    \resizebox{\linewidth}{!}{
    \begin{tabular}{l|lllll|lllllll}
    \toprule
    Datasets       & \multicolumn{5}{c}{Office-Caltech10}         &   \multicolumn{7}{c}{DomainNet}                                                        \\ \cline{2-13}
        Domains & A & C & D & W & Avg & C & I & P & Q & R & S & Avg. \\
        \midrule
        CLIP \cite{radford2021learning} & 19.40  & 18.32  & 21.87  & 18.59  & 19.55  & 49.89 & 47.23 & 53.61 & 32.10 & 48.19 & 50.79 & 46.96 \\ 
        CoOp \cite{zhou2022learning} & 41.54  & 15.55  & 56.04  & 43.60  & 39.18  & 83.42 & 53.28 & 80.80 & 49.41 & 75.18 & 82.88 & 70.83 \\ 
        PromptFL \cite{guo2023promptfl} & 96.34  & 91.57  & 97.96  & 98.30  & 96.04  & 95.28 & 73.72 & 94.50 & 61.60 & 95.72 & 95.43 & 86.04 \\ 
        Prompt+Prox \cite{li2020federated} & 96.13  & \textbf{92.52}  & 97.57  & 97.96  & 96.05  & 95.47 & 69.44 & 94.95  & 61.24 & 75.18 & 95.41 & 81.95 \\ 
        FedOTP \cite{li2024global}& 95.88& 92.13 & \textbf{99.15} & 97.15 & 96.07 & 94.10 & 70.57 & 89.88& 55.80 & 94.93& 92.73& 83.00\\
        \midrule
        FedPGP & \textbf{96.55}  & 91.92  & 98.93  & \textbf{98.75}  & \textbf{96.54} & \textbf{96.45} & \textbf{74.46} & \textbf{95.43} & \textbf{62.12} & \textbf{96.06} & \textbf{96.05} & \textbf{86.76} \\ 
        \bottomrule
    \end{tabular}
    }
\end{table*}

\begin{table*}[!ht]
    \centering 
    \renewcommand\arraystretch{1.12}
    \caption{Accuracy comparison (\%) on the Pathological Non-IID setting over 10 clients.}
    \label{acc}
    \resizebox{0.88\linewidth}{!}{
    \begin{tabular}{l | ccccc}
    \toprule
        Methods  &OxfordPets & Flowers102 & DTD & Caltech101 & Food101 \\
            \midrule
        CoOp \cite{zhou2022learning} & 83.21±1.30 & 70.14±0.76 & 44.23±0.63 & 87.37±0.44 & 70.43±2.42 \\ 
        PromptFL \cite{zhou2022learning} & 90.79±0.61 & 72.80±1.14 & 54.11±0.22 & 89.70±1.99 & 77.31±1.64 \\ 
        PromptFL+FT \cite{cheng2021fine} & 91.23±0.50 & 72.31±0.91 & 53.74±1.36 & 89.70±0.25 & 77.16±1.56 \\ 
        Prompt+PER \cite{arivazhagan2019federated} & 89.50±1.62 & 72.11±1.35 & 50.23±0.82 & 86.72±1.45 & 71.29±1.87 \\ 
        Prompt+Prox \cite{li2020federated} & 89.24±0.41 & 66.40±0.29 & 44.26±1.11 & 89.41±0.55 & 76.24±1.94 \\ 
        Prompt+AMP \cite{huang2021personalized} & 80.21±0.44 & 69.10±0.13 & 47.16±0.92 & 87.31±1.60 & 74.48±1.71 \\ 
        pFedPrompt \cite{guo2023pfedprompt} & 91.84±0.41 & 86.46±0.15 & 77.14±0.09 & 96.54±1.31 & 92.26±1.34 \\ 
        \midrule
        FedPGP & \textbf{98.96±0.42} & \textbf{99.29±0.03} & \textbf{91.52±0.41} & \textbf{98.90±0.19} & \textbf{95.52±0.15} \\
        \bottomrule
    \end{tabular}
    }
\end{table*}

\noindent \textbf{Implementation Details.} 
All methods presented in this paper are based on a frozen CLIP using two backbones, ResNet50 \cite{he2016deep} and ViT-B16 \cite{dosovitskiy2020image}, defaulting to ViT-B16 if not explicitly specified. In federated learning, we set the client's local training epoch $E = 1$ and communication round $T=150$ with $N=100$ clients and partition rate $r=10\%$ for CIFAR-10/CIFAR-100 datasets. Besides, we consider training epoch $E = 2$ and communication round $T=25$ with client numbers $N=10$ and a full partition rate, i.e., $r=100\%$ for other datasets. The low-rank decomposition bottleneck is set to $b=8$, and the hyperparameter $\mu$ for the contrastive loss is set to $1$. We employ cosine similarity as the metric function in contrastive loss. For the setting of learnable prompts, the length of prompt vectors $p$ is $16$ with a dimension of $512$, token position is ``end" with ``random" initialization. Apart from the few-shot learning, batch sizes are set to $32$ during training and $100$ during testing.  Additional implementation details can be found in the Appendix Section \ref{experimental_setup}.

\subsection{Performance Evaluation}

\noindent \textbf{Base-to-Novel Class Generalization.} 
We evaluated the performance of FedPGP against baselines on their local classes, base classes, and novel classes respectively. We present the harmonic mean (HM) of these three accuracies to demonstrate the overall performance. The experiment results are summarized in Table \ref{basetonew}. As indicated in Table \ref{basetonew}(a), FedPGP achieves the best performances in local classes, highlighting its exceptional personalization capability. Moreover, FedPGP outperforms other methods in both base classes and the harmonic meanwhile also exhibiting the second-best performance in novel classes. These results show its exceptional capacity for balancing personalization and generalization. CoOp achieves the second-best performance in local classes owing to the pathological setting. However, it cannot effectively generalize its performance to other base classes and new classes. When it comes to FPL, PromptFL and PromptProx sacrifice the personalization capability in local classes to gain better generalization ability.

\noindent \textbf{Leave-One-Domain-Out Generalization.} 
Table \ref{leave_one_domain_out} shows the average classification accuracy with leave-one-domain-out validation on Office-Caltech10 and DomainNet. As we can see, FedPGP achieves the highest average accuracy and outperforms all baselines across nearly all target domains. Through local prompt tuning, CoOp's domain generalization capabilities generally surpass those of CLIP.
We notice that FPL enhances the model's domain generalization power, marking a significant improvement compared to the local approach. Moreover, FedPGP enhances the model's ability to generalize while accomplishing personalization, which proves the effectiveness of our framework's design. We provide the detailed classification accuracy on each source domain within the Office-Caltech10 dataset in Table \ref{office_detail}. Additional experiment results on specific client domain generalization are available in the Appendix Section \ref{appendix_domain}.

\begin{table*}[!ht]
\renewcommand\arraystretch{1.1}
    \centering
    \caption{The detailed classification accuracy using leave-one-domain-out validation on Offica-Caltech10 dataset.}
    \label{office_detail}
    \resizebox{0.8\linewidth}{!}{
    \begin{tabular}{l|c|cccc|c}
\toprule
Datasets                & \multicolumn{6}{c}{Office-Caltech10}      \\ \cline{2-7} 
Source Domains            &     & Amazon & Caltech & DSLR  & Webcam & Avg.  \\ \midrule
\multirow{4}{*}{CoOp \cite{zhou2022learning}}   
      & Amazon         & ——     & 89.03   & 16.49 & 19.1   & 41.54±41.15 \\
      & Caltech         & 26.89  & ——      & 5.87  & 13.89  & 15.55±10.61 \\
      & DSLR         & 64.96  & 86.62   & ——    & 16.56  & 56.04±35.87 \\
      & Webcam            & 50.16  & 76.94   & 3.72  & ——     & 43.6±37.05  \\ \midrule
\multirow{4}{*}{FedPGP}  & Amazon  & ——     & 96.45   & 96.03 & 97.18  & \textbf{96.55±0.58} \\
      & Caltech          & 94.66  & ——      & 86.92 & 93.59  & \textbf{91.92±4.19} \\
      & DSLR          & 98.08  & 99.36   & ——    & 99.36  & \textbf{98.93±0.74} \\
      & Webcam            & 98.98  & 98.98   & 98.3  & ——     & \textbf{98.75±0.39} \\ 
\bottomrule
\end{tabular}
    }
\end{table*}

\noindent \textbf{Evaluation on Personalization.} 
We report the performance of FedPGP against baselines in Table \ref{acc}, \ref{cifar10}. To facilitate comparison, we present the results in Table \ref{acc} utilizing ResNet50 as the backbone, aligning with the setting in \cite{guo2023pfedprompt}. As shown in Table \ref{acc}, our FedPGP demonstrates significantly superior performance compared to baseline methods across all datasets. This confirms that our framework's ability to personalize effectively is successful in addressing extreme non-IID scenarios. Table \ref{cifar10} shows the results of FedPGP and baseline methods on CIFAR-10 and CIFAR100 datasets with Dirichlet Non-IID setting over 100 clients with 10\% partition. In the scenario with Dirichlet settings and the substantial number of clients, our approach FedPGP consistently demonstrates superior performance compared to the baseline methods. This further emphasizes the effectiveness of our approach.
\begin{table}[]
\caption{Accuracy comparison (\%) on the Dirichlet Non-IID setting in CIFAR-10 and CIFAR-100 over 100 clients.}
\resizebox{\linewidth}{!}{
\label{cifar10}
\renewcommand\arraystretch{1.1}
\begin{tabular}{l|cc}
\toprule
Methods    & CIFAR-10   & CIFAR-100   \\
\midrule
CLIP \cite{radford2021learning}       &   87.52$\pm$0.56    &   64.83$\pm$0.49          \\
CoOp \cite{zhou2022learning}      &   93.13$\pm$0.34      &      74.78$\pm$0.41  \\
PromptFL \cite{zhou2022learning}  &   92.32$\pm$0.79    &   73.72$\pm$0.61      \\
Prompt+Prox \cite{li2020federated} &   91.79$\pm$0.46       &  71.08$\pm$0.89     \\
\midrule
FedPGP     &   \textbf{94.82$\pm$0.37}          &     \textbf{77.44$\pm$0.15}          \\
\bottomrule
\end{tabular}
}
\end{table}

\subsection{Ablation Study}

\begin{figure*}[t]
\centering
\includegraphics[width=1\textwidth]{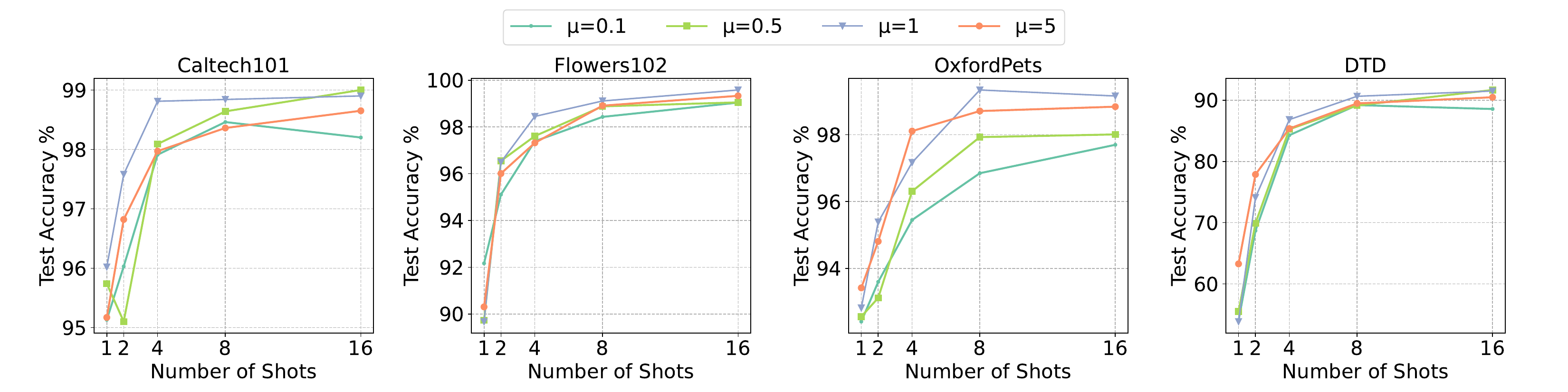}
\caption{Quantitative comparisons on four datasets across varying shot numbers and parameter $\mu$ of contrastive loss in FedPGP over 10 clients.} 
\label{ablation_mu}  
\end{figure*}

\textbf{Effect of Parameter $\mu$ of Contrastive Loss}
In this subsection, we investigated the impact of the contrastive loss parameter $\mu$ in a Pathological Non-IID setting across four datasets with varying shot numbers. The results are presented in Figure \ref{ablation_mu}, which shows an improvement in test accuracy with an increase in the number of shots. 
Upon observation, optimal results are mostly achieved with $\mu$ set to 1 in experiments, leading to our adoption of $\mu=1$ for other experiments.

\textbf{Effective of Low-rank Adaption}
In this subsection, we explored the effectiveness of low-rank adaptation by comparing it with full-rank adaptation. The results of the two methods are shown in Table \ref{ablationgen}. As we can see, full-rank adaptation achieves the best performance on local classes but it completely overwrites the global prompt, resulting in a loss of category-
agnostic knowledge and the generalization capacity. Although low-rank adaptation performance on the local class is below the full-rank adaptation, it significantly outperforms the full-rank adaptation in terms of base-to-novel generalization.

\begin{table}[t]
\centering
\caption{Accuracy (\%) of ablation study on adaption and additional loss for clients’ local classes and Base-to-novel generalization.}
\renewcommand\arraystretch{1.2}
\resizebox{\linewidth}{!}{
\label{ablationgen}
\begin{tabular}{l|ccc|c}
\toprule
     Methods      & Local &  Base & Novel & HM    \\
\midrule
FedPGP w/o Positive           &94.63 & 84.68 & 77.75 & 85.13     \\
FedPGP w/ Full-rank Adaption  & \textbf{ 98.57}       & 48.00       & 63.40 & 64.17 \\
FedPGP  & 95.67       & \textbf{85.69}       & \textbf{81.75} & \textbf{87.33} \\
\bottomrule
\end{tabular}
}
\end{table}

\textbf{Effective of Contrastive Loss} 
We demonstrate the efficacy of contrastive loss in achieving balance by separately testing the generalization ability of the model without positive pairs and the personalization ability of the model with negative pairs. Table \ref{ablationgen} shows the performance of the model without knowledge-guidance from CLIP (positive pairs), compared to the performance when the contrastive loss is employed. FedPGP outperforms the model without knowledge-guidance across all three accuracies and the harmonic mean. Table \ref{ablationacc} shows the performance of the model without pushing the representation of global prompt and personalized prompt (negative pairs), compared to the performance when the contrastive loss is employed. The results show the negative pairs enhance the model's ability to personalize for Non-IID data distribution in federated prompt learning. In general, our ablation study on contrastive loss demonstrates its ability to balance personalization and generalization in federated prompt learning.
\begin{table}[t]
\centering
\setlength\tabcolsep{2pt}
\renewcommand\arraystretch{1}
\caption{Accuary (\%) of ablation study on additional loss for personalization.}
\resizebox{\linewidth}{!}{
\label{ablationacc}
\begin{tabular}{l|ccccc}
\toprule
Methods       & OxfordPets  & Flowers102   & DTD      & Caltech101     & Food101     \\
\midrule
FedPGP w/o Negative       & 97.65±0.20   & 98.63±0.11   & 90.78±0.31      & 98.48±0.17    & 94.72±0.18                 \\
FedPGP & \textbf{98.96±0.42}           & \textbf{99.29±0.03} & \textbf{91.52±0.41 }          & \textbf{98.90±0.19 }          & \textbf{95.52±0.15  }         \\
\bottomrule
\end{tabular}
}

\end{table}


\section{Conclusion}
In this paper, we propose a novel approach named FedPGP, which represents a pioneering effort to harmonize personalization and generalization in federated prompt learning. In our approach, each client gains generalization capabilities through knowledge-guidance of CLIP and acquires personalization abilities by adapting the global prompt to a personalized prompt. Further, with low-rank decomposition adaptation and an extra contrastive loss, FedPGP learns a personalized prompt for each client in heterogeneous federated scenarios while preserving the remarkable generalization capacity in pre-trained Vision-Language models. 
Extensive experiments on various datasets explored base-to-novel generalization in both unseen categories and domains, showing the superiority of FedPGP in balancing generalization and personalization.
In future work, we aim to explore the theoretical foundations of low-rank adaptation in federated prompt learning.

\section*{Acknowledgement}
 This work was supported by NSFC (No.62303319), Shanghai Sailing Program (22YF1428800, 21YF1429400), Shanghai Local College Capacity Building Program (23010503100), Shanghai Frontiers Science Center of Human-centered Artificial Intelligence (ShangHAI), MoE
 Key Laboratory of Intelligent Perception and Human Machine Collaboration (ShanghaiTech University), and Shanghai Engineering Research Center of Intelligent Vision and Imaging. 
\clearpage

\section*{Impact Statement}
There are many potential societal consequences of our work, none of which we feel must be specifically highlighted here. 
\bibliography{icml2024/ref}
\bibliographystyle{icml2024}

\newpage
\appendix
\onecolumn

\section{Experimental Details}

\subsection{Dataset Setup}
\label{dataset_setup}
\begin{figure*}[htbp]
  \centering
  \subfigure[DomainNet]{\includegraphics[width=0.46\textwidth]{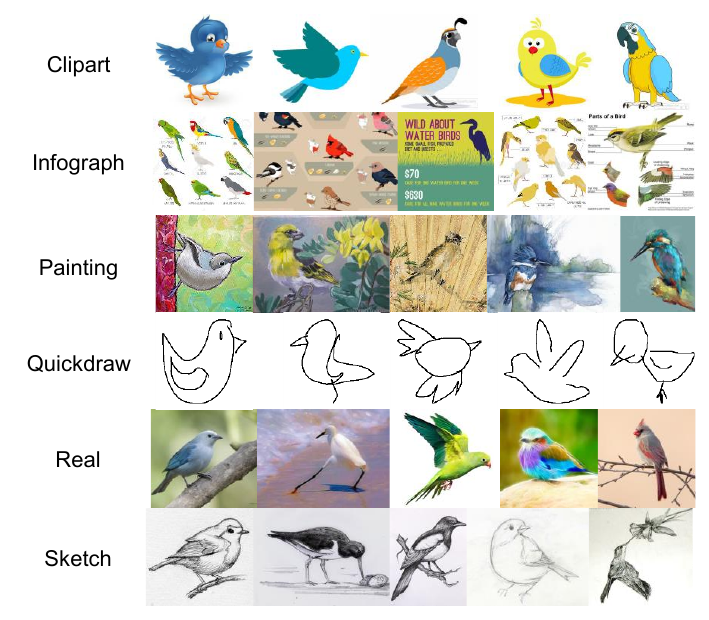}}
  \subfigure[Office-Caltech10]{\includegraphics[width=0.49\textwidth]{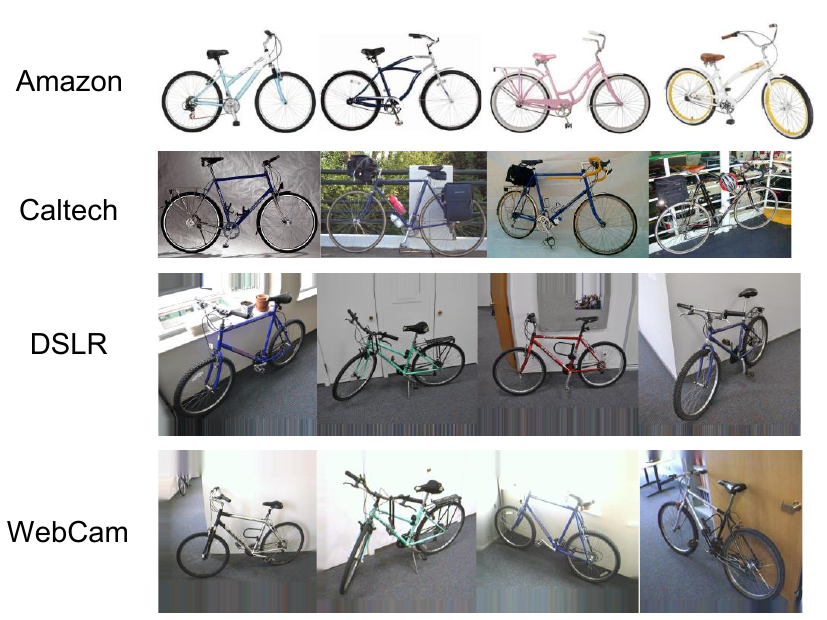}}
  \caption{Visual examples of raw instances from two datasets with multiple domains: ``Bird" in DomainNet (left) and ``Bike" in Office-Caltech10 (right).}
  \label{example_feature_shift}
\end{figure*}

For our evaluation, we've chosen nine diverse visual classification datasets as our benchmark. Table \ref{datasets_details} provides a detailed overview, including information on original tasks, class numbers, training and testing sample sizes, and domain counts.
In datasets with multiple domains, we utilize the well-established Office-Caltech10 benchmark, featuring four domains: Amazon, Caltech, DSLR, and WebCam. These domains capture variations arising from different camera devices and real-world environments. Additionally, we leverage DomainNet, a large-scale dataset comprising six domains: Clipart, Infograph, Painting, Quickdraw, Real, and Sketch. We focus on training with 10 selected classes from each dataset. Visual examples of raw instances from these two multi-domain datasets can be found in Figure \ref{example_feature_shift}. 

\begin{table*}[!ht]
\renewcommand\arraystretch{1.2}
\centering
\caption{Statistical details of datasets used in experiments.}
\label{datasets_details}
\resizebox{0.98\textwidth}{!}{
\begin{tabular}{l|ccccl}
\toprule
Dataset  & Classes & Train & Test & Domains & Task\\ \midrule
OxfordPets \cite{parkhi2012cats}                & 37                         & 2,944                         & 3,669            &1   & Fine-grained pets recognition \\
Flowers102 \cite{nilsback2008automated}                       & 102                  & 4,093                            & 2,463      &1   & Fine-grained flowers recognition   \\
DTD \cite{cimpoi2014describing}               & 47                         & 2,820                       & 1,692              &1  & Texture recognition   \\ 
Caltech101 \cite{fei2004learning}                         & 100                   & 4,128                            & 2,465        &1  & Object recognition \\
Food101 \cite{bossard2014food}           &101    &50,500   &30,300        &1   & Fine-grained food recognition \\  \midrule
CIFAR10 \cite{krizhevsky2009learning}               & 10                        & 50,000                        & 10,000         &1   & Image Classification   \\
CIFAR100 \cite{krizhevsky2009learning}              & 100                         & 50,000                        & 10,000         &1    & Image Classification   \\ \midrule
DomainNet \cite{peng2019moment}                  & 10                        & 18278                        & 4573              &6  &  Image recognition   \\ 
Office-Caltech10 \cite{gong2012geodesic}                & 10                        & 2025                        & 508          &4   &  Image recognition  \\
\bottomrule
\end{tabular}
}
\end{table*}


\subsection{Experimental Setup}
\label{experimental_setup}
We employ SGD optimizer with learning rate $\eta=0.001$. The experiments were conducted three times using different seeds. We calculated the average performance and the final result in federated prompt learning is obtained by averaging the performance across all clients. All experiments are conducted with Pytorch on NVIDIA A40 GPUs.

\noindent \textbf{Base-to-Novel Class Generalization.} 
For Base-to-Novel generalization, we separate each dataset into base and novel classes equally and distribute the base classes to each client without overlapping. Each client trains their local model on their local classes, and we evaluate their personalized prompt on both local classes, base classes (classes seen on other clients but unseen during local training), and novel classes (unseen in the whole training process). The accuracy is the average overall 10 clients.

\noindent \textbf{Leave-One-Domain-Out Generalization.} 
For Leave-One-Domain-Out generalization, each client participating in the federated learning system is assigned data from one of the distinct domains. We pick one domain to serve as the target domain and use the rest of the domains as source domains. Each client possesses a distinct source domain for training and then tests its model generalization ability on the whole target domain. The accuracy is the average overall 3 clients in Office-Caltech10 and 5 clients in DomainNet.

\noindent \textbf{Personalization.}
 For evaluation of personalization, we apply Dirichlet distribution on CIFAR-10 and CIFAR-100 over 100 clients. Specifically, the datasets are partitioned randomly among clients using a symmetric Dirichlet distribution with hyperparameter $\alpha$ = 0.3. Besides, we employ the Pathlogicacl setting the same as in base-to-novel class generalization with non-overlapping classes across 10 clients for other datasets. 
 
 \noindent \textbf{Effect of Individual Components.}
 For the ablation study, we employ $\mathcal{L}_{neg} = 1 - \text{sim}(z_G,z_i)$ as the additional loss for FedPGP without positive pairs, and $\mathcal{L}_{pos}= \text{sim}(z_G,z_C)$ as the additional loss for FedPGP without negative pairs. The hyperparameter is set as $\mu = 1$ for the additional loss.

\section{Additional Experimental Results}


\subsection{Detailed Results of Leave-One-Domain-Out Generalization}
\label{appendix_domain}
In Table \ref{office_detail} and Table \ref{domainnet_detail}, we provide the detailed classification accuracy on each source domain within Office-Caltech10 and DomainNet datasets, respectively. Notably, as PromptFL and PromptProx utilize a single global prompt, their results remain consistent across different source domains. Therefore, the presented results specifically focus on CoOp and our FedPGP, both employing distinct local models for each client. To be specific, the values shown in the table indicate the testing results on the target domain across clients with different source domains. Comparing the results of CoOp and our FedPGP, we observe that FedPGP consistently outperforms CoOp in all cases with a significantly smaller standard deviation, showcasing the robust generalization capability of our proposed method.

\begin{table*}[!ht]
\renewcommand\arraystretch{1.1}
    \centering
    \caption{The detailed classification accuracy using leave-one-domain-out validation on DomainNet dataset.}
    \label{domainnet_detail}
    \resizebox{1\linewidth}{!}{
\begin{tabular}{l|c|cccccc|c}
\toprule
Datasets                & \multicolumn{7}{c}{DomainNet}                                                                              \\ \cline{2-9} 
Source Domains     &            & Clipart & Infograph & Painting & Quickdraw & Real  & Sketch & Avg. \\ \midrule
\multirow{6}{*}{CoOp \cite{zhou2022learning}}   
        & Clipart       & ——      & 81.17     & 82.80    & 69.98     & 90.71 & 92.46                      & 83.42±8.96                    \\
        & Infograph     & 57.41   & ——        & 60.00    & 30.64     & 65.09 & 53.24                      & 53.28±15.20                   \\
        & Painting      & 86.80   & 70.88     & ——       & 67.06     & 92.62 & 83.04                      & 80.80±11.67                    \\
        & Quickdraw     & 48.50   & 44.02     & 50.00    & ——        & 53.02 & 51.50                      & 49.41±3.94                    \\
        & Real          & 89.95   & 79.27     & 93.70    & 27.29     & ——    & 85.69                      & 75.18±30.05                    \\
        & Sketch        & 86.55   & 84.27     & 86.91    & 64.51     & 92.15 & ——                         & 82.88±12.09                    \\ \midrule
\multirow{6}{*}{FedPGP} & Clipart & ——      & 96.23     & 96.80    & 95.96     & 96.27 & 96.99                      & \textbf{96.45±0.43}                    \\
        & Infograph     & 76.45   & ——        & 74.44    & 69.29     & 76.24 & 75.90                      & \textbf{74.46±3.21}                    \\
        & Painting      & 96.18   & 95.89     & ——       & 93.01     & 96.02 & 96.08                      & \textbf{95.43±1.50}                    \\
        & Quickdraw     & 66.22   & 52.30     & 63.42    & ——        & 62.30 & 66.36                      & \textbf{62.12±6.11}                    \\
        & Real          & 97.12   & 95.75     & 97.03    & 93.17     & ——    & 97.25                      & \textbf{96.06±1.87}                    \\
        & Sketch        & 96.71   & 95.88     & 96.56    & 94.75     & 96.38 & ——                         & \textbf{96.05±0.81}                    \\ 
\bottomrule
\end{tabular}
    }
\end{table*}

\subsection{Detailed Results of Individual Components in Base-to-Novel Generalization}
Table \ref{ablationgen_detail} presents the per-dataset results for each component of our FedPGP framework in the Base-to-novel generalization setting. The results demonstrate the effectiveness of CLIP knowledge-guidance in enhancing performance for both base and novel classes.  Additionally, even though full-rank adaptation outperforms our low-rank adaptation on local classes, its generalization on both base and novel classes significantly diminishes due to overwriting the global prompt.  These findings emphasize the efficacy of FedPGP in enhancing model generalization across diverse datasets. 

\begin{table}[h!]
\centering
\renewcommand\arraystretch{1.1}
\caption{Accuracy (\%) of ablation study on adaption and additional loss for clients’ local classes and Base-to-novel generalization.}
\resizebox{0.75\linewidth}{!}{
\label{ablationgen_detail}
\begin{tabular}{ll|ccc|c}
\toprule
Dataset            & Methods   & Local & Base  & Novel & HM    \\ 
\midrule
\multirow{3}{*}{\begin{tabular}[c]{@{}l@{}}Average over \\ 5 datasets\end{tabular}} 
& FedPGP w/o Positive          & 94.63 & 84.68 & 77.75 & 85.13 \\
& FedPGP w/ Full-rank Adaption & \textbf{98.57} & 48.00 & 63.40 & 64.17 \\
& FedPGP                       & 95.67 & \textbf{85.69} & \textbf{81.75} & \textbf{87.33} \\ 
\midrule
\multirow{3}{*}{Oxford Pets} 
& FedPGP w/o Positive          & 95.88 & 95.44 & 96.77 & 96.03 \\
& FedPGP w/ Full-rank Adaption & \textbf{99.94} & 41.28 & 73.32 & 62.67 \\
& FedPGP                       & 96.65 & \textbf{95.87} & \textbf{97.33} & \textbf{96.61} \\
\midrule
\multirow{3}{*}{Flowers102} 
& FedPGP w/o Positive          & 98.73 & 77.18 & 62.22 & 76.61 \\
& FedPGP w/ Full-rank Adaption & \textbf{99.91} & 35.43 & 53.97 & 52.85 \\
& FedPGP                       & 99.68 & \textbf{78.48} & \textbf{75.11} & \textbf{83.13}   \\ 
\midrule
\multirow{3}{*}{DTD} 
& FedPGP w/o Positive          & 87.34 & 67.33 & 49.83 & 64.70 \\
& FedPGP w/ Full-rank Adaption & \textbf{96.29} & 25.12 & 34.81 & 38.01 \\
& FedPGP                       & 89.07 & \textbf{69.65} & \textbf{54.25} & \textbf{68.15}   \\ 
\midrule
\multirow{3}{*}{Caltech101} 
& FedPGP w/o Positive          & 98.34 & 96.02 & 92.57 & 95.58 \\
& FedPGP w/ Full-rank Adaption & \textbf{99.89} & 75.94 & 81.08 & 84.48 \\
& FedPGP                       & 99.46 & \textbf{96.09} & \textbf{93.62} & \textbf{96.33} \\ 
\midrule
\multirow{3}{*}{Food101} 
& FedPGP w/o Positive          & 92.85 & 87.42 & 87.36 & 89.14 \\
& FedPGP w/ Full-rank Adaption & \textbf{96.84} & 62.21 & 73.8 & 75.09 \\
& FedPGP                       & 93.51 & \textbf{88.37} & \textbf{88.44} & \textbf{90.04}   \\ 
\bottomrule
\end{tabular}
}
\end{table}

\subsection{Effect of Number of Bottleneck}
In this subsection, we explore the impact of the number of bottleneck $b$ in our low-rank decomposition of adaptation term $\Delta p_i$. We present the accuracy results considering the impact of both the bottleneck and shot number using a random seed. It can be observed that the classification accuracy improves as the bottleneck and shot number increase, showing the number of bottleneck determines the extent to which the knowledge in the global prompt is rewritten. We select the number of bottleneck $b=8$ for the balance of generalization and personalization.
\begin{table}[h!]
\centering
\renewcommand\arraystretch{1.1}
\caption{Quantitative comparisons on 4 datasets across varying number of shots with different number of bottleneck in FedPGP over 10 clients.}
\resizebox{0.72\linewidth}{!}{
\label{ablationbottleneck}
\begin{tabular}{lcccccc}
\toprule
Dataset                      & Bottleneck & 1 shot & 2 shots & 4 shots & 8 shots & 16 shots \\ 
\midrule
\multirow{4}{*}{Oxford Pets} & 1   &  92.4	&92.89 &	93.96&	94.28&	95.12     \\
                             & 2  & 92.39&	93.04	&94.93	&95.91	&96.39
      \\
                             & 4   &  92.51&	\textbf{93.62}&	94.66&	96.72&	97.32
       \\
                             & 8 &  \textbf{93.16}	&93.12&	\textbf{96.31}	&\textbf{97.93}	&\textbf{97.81}
       \\ 
                             \midrule
\multirow{4}{*}{Flowers102} & 1 &  86.89&	91.92&96.26&	98.56	&98.75
         \\
                             & 2  & 87.79	&93.95&	96.28	&97.60	&98.71
        \\
                             & 4 &  87.77&	94.86&	97.61&	\textbf{98.92}	&\textbf{99.37}
        \\
                             & 8 &  \textbf{89.74}	&\textbf{96.55}&	\textbf{97.64}	&98.88	&99.05
        \\
                             \midrule
\multirow{4}{*}{DTD}     & 1 & 53.13	&60.52&	70.41&	85.61	&83.00
         \\
                             & 2 &  52.63	&58.77	&73.97	&87.75	&91.05
        \\
                             & 4  &  55.02	&66.05	&76.80	&\textbf{89.27}&	90.08
         \\
                             & 8& \textbf{55.47}&	\textbf{69.91}	&\textbf{85.27}&	89.16&	\textbf{92.00}
         \\
                             \midrule
\multirow{4}{*}{Caltech101} & 1 &  93.46	&93.93&	96.06	&97.62	&98.40
       \\
                             & 2 & 93.27	&94.36	&96.69&	97.89	& 98.33
         \\
                             & 4  &  94.44	&\textbf{96.32}	&97.02&	98.20&	98.30
       \\
                             & 8  & \textbf{95.74}&	95.10&	\textbf{98.09}&	\textbf{98.28}	&\textbf{99.00}
         \\
\bottomrule
\end{tabular}
}
\end{table}

\subsection{Learning Curves}
\begin{figure}[h]
\centering
\includegraphics[width=1\textwidth]{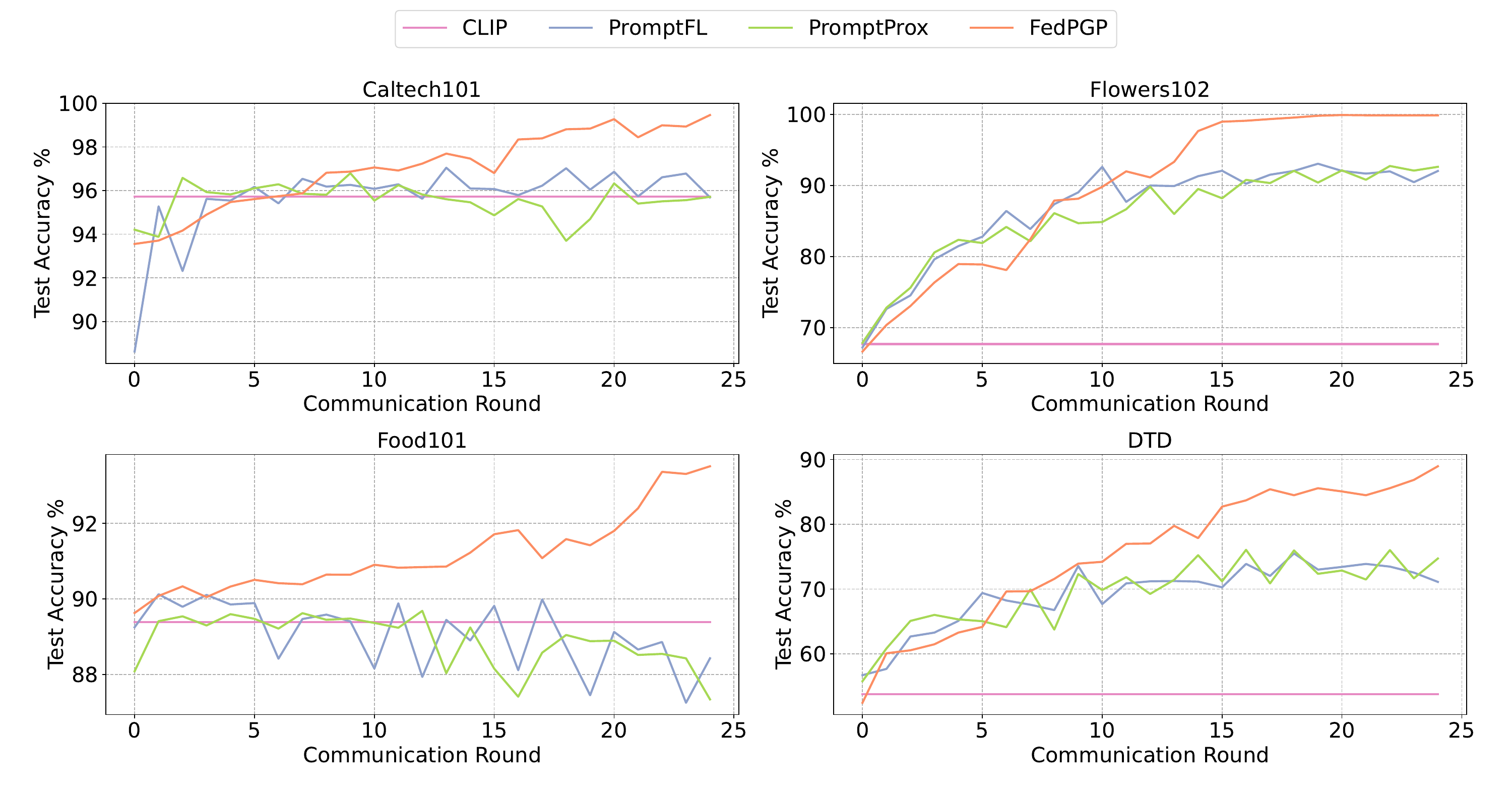}
\caption{Accuracy learning curves of FedPGP and baselines on four datasets over 10 clients.} 
\label{learning_curve}  
\end{figure}

To analyze the convergence pattern of our FedPGP, we visualized the test accuracy across 10 clients with a local training epoch $E=2$ and communication round $T=25$. The results are illustrated in Figure \ref{learning_curve}, revealing accelerated convergence and enhanced stability exhibited by FedPGP.

\end{document}